# A Causal-based Framework for Multimodal Multivariate Time Series Validation Enhanced by Unsupervised Deep Learning as an Enabler for Industry 4.0

Cedric Schockaert
Paul Wurth S.A.
Department of Process Automation
Luxembourg, Luxembourg
cedric.schockaert@paulwurth.com

*Abstract*—An advanced conceptual validation framework for multimodal multivariate time series defines a multi-level contextual anomaly detection ranging from an univariate context definition, to a multimodal abstract context representation learnt by an Autoencoder from heterogeneous data (images, time series, sounds, etc.) associated to an industrial process. Each level of the framework is either applicable to historical data and/or live data. The ultimate level is based on causal discovery to identify causal relations in observational data in order to exclude biased data to train machine learning models and provide means to the domain expert to discover unknown causal relations in the underlying process represented by the data sample. A Long Short-Term Memory Autoencoder is successfully evaluated on multivariate time series to validate the learnt representation of abstract contexts associated to multiple assets of a blast furnace. A research roadmap is identified to combine causal discovery and representation learning as an enabler for unsupervised Root Cause Analysis applied to the process industry.

## I. INTRODUCTION AND BACKGOUND

In the current era of digitalization and optimization of industrial processes, large number of sensors are being installed aiming at measuring and storing enormous quantities of data. This data is bringing means to control the underlying process by the definition of rules, and to train data-driven models for multiple purposes such as process forecasting, predictive/prescriptive maintenance and to recommend the best appropriate actions to the process engineer for optimal process regulation. The basement of a successful automation system relying on advanced data analytics is enabled by implementing a data validation strategy. A data validation framework is proposed in this article and is a generalization of the validation pyramid we presented in the context of the blast furnace [1]. It covers, at the foundation level, univariate time series methods for detection of anomalies, and is extended by learning contextual representation from multimodal time series to reach the ultimate representation of the complete end-to-end process. To that end, unsupervised deep learning approaches are the core technologies elevating the validation to a level beyond the capability of human by learning causal relations between enormous quantities of signals measured by sensors to characterize an industrial process. The data validation framework should include an additional level of validation dedicated to data targeting a training of a data-driven model in order to detect not solely anomalies but to ensure the data to be unbiased, statistically representative of the underlying process by validating the causal relations by means of advanced analytics, and verifying data assumptions by hypothesis testing when applicable.

To enable the automation of an industrial system, it is often required to record temporal data of multiple formats: images, multivariate time series, sounds, behavioral data, text or comments by process engineers. In order to globally learn from this large and various quantity of data, dedicated architectures must be researched aiming at extracting and correlating events to reach a global understanding of the process leading to improved context representation learning by an unsupervised approach, or better forecasting power of a supervised model. Deep learning has been vastly applied on images to extract meaningful features in a latent space. Other specific architectures such as Long Short-Term Memory (LSTM) are recurrent neural networks developed to generate abstract features from multivariate time series data by solving inherent limitation of recurrent neural networks in relation with the vanishing gradient limiting the size of the temporal window for learning representations. Natural Language Processing is extracting relevant knowledge from unstructured text that can be used as new features (ex: sentiment analysis) for training a multimodal machine learning model. The extraction of the information or its transformation into meaningful insight is bringing unsupervised deep learning into one of the most promising approach for the next generation of Artificial Intelligence (AI). Multimodal unsupervised learning allows understanding the world, and therefore the context, similarly to a child learning to walk using his five basic senses. An impressive move towards unsupervised learning is currently happening in the research community accepting the limitation of existing labels for supervised learning, due to their bias without a proper causal-based validation framework, potential low quality and cost for generation. Representation learning and particularly out-of-distribution adaptation has been emphasized at the annual NeurIPS conference in Toronto in 2019 by Prof Y. Bengio [2] as being a research field on which the research community must focus for the development of new AI generation embedding the consciousness of the surrounding environment learned with unsupervised approaches. Learning causal relations defining the environment, temporally and spatially, is a trigger for the process engineer in the industry to learn from his data. Root cause analysis, as an applicative example of unsupervised approach, brings new insights about the process captured by a large amount of sensors. It has been very clearly identified that process engineers are not always ready to rely on machine learning models to control and optimize their respective processes, or to predict the occurrence of a failure in the future (Remaining Useful Life of assets). The priority is the understanding of the current process, to judge from thousands of measurements with minute granularity if the actions he has triggered are leading the process to reach the defined target, but also the quantification of unexpected outcomes. The application of machine learning in the industry is particularly challenging, as often the main user of a black-box data-driven model is the process expert himself with sometimes years of experience. In order to reach the acceptance level of a prediction generated by a black-box data-driven model, it is particularly crucial to prioritize the research for training models helping the industry to reach a better level of understanding of their respective processes. This leads to the need to recognize if an anomaly is originating from a faulty sensor or induced by rare process phenomena. Detecting a deviation from normal relations between sensors signals is not enough to associate that deviation to a faulty sensor. However, root cause analysis brings a justification of a signal deviation to the process engineer enabling a judgment about the origin of that deviation being sensor or process related.

Few methods implementing an end-to-end data validation are emerging from the scientific research. Often approaches are based on the validation of data in a limited context such as univariate time series, and complex anomalies stay undiscovered. In [3] they apply machine learning methods for massive image data validation and selection for training a data-driven model. More recently, in [4], they proposed a conceptual data validation approach based on a risk calculation of poor data quality for the machine learning model to train. In [5] software engineers are defining data quality checks and associated thresholds, however they are facing the challenge of tuning those thresholds for optimal anomaly classification, as first they are not the domain expert, and secondly they are having limited representation of the context even not captured by a dedicated unsupervised deep leaning model for contextual representation learning. The nature of the data being dynamic (either qualitatively or quantitatively), heterogeneous (text, image, tabular files) and its storage often silo-based and delivered either in batches or in real-time mode, increases further the risk of a technical debt in the absence of the guidance offered by an automatic framework for advanced data validation [6].

The framework for multi-level data validation proposed in this article is relying on the knowledge of the domain expert to define basic thresholds such as min/max values for each signals, and to use simulation models for further cross-validation with sensor signals. The proposed solution is implementing a dedicated level for advanced contextual representation learning by an unsupervised deep learning model to bring automation of anomaly detection without the need to define thresholds that is simply not feasible for a human facing several thousands of heterogeneous data with complex spatio-temporal causal relations and potential unknown confounders inducing false correlations between variables. Furthermore, the last level is dedicated to the causal discovery in a training dataset allowing the domain expert to assess any potential bias, to learn unknown relations in his process, allowing him to optimize his process by communicating with the machine and therefore enabling an AI culture change in the process industry.

## II. DESCRIPTION OF THE PROPOSED FRAMEWORK

As an answer to the requirement to validate multimodal dataset characterizing industrial processes, a multi-level data validation framework is elaborated and presented in Fig. 1, where each level of the framework is dedicated to a specific validation purpose with an increased level of abstraction to represent the context until learning the causal relations of the underlying process from observational data. The application of individual level is either *online* by providing a validation for any live inputs, or *offline* for the validation of a historical dataset. Causal discovery is allowing the process engineer to get insights about his process and is contributing therefore to the acceptance of AI. However, an abstract representation in a latent space of the process normality brings significant contextual deep features to train a black-box data-driven model but limited additional values to the expert without the explanation given by a causal model.

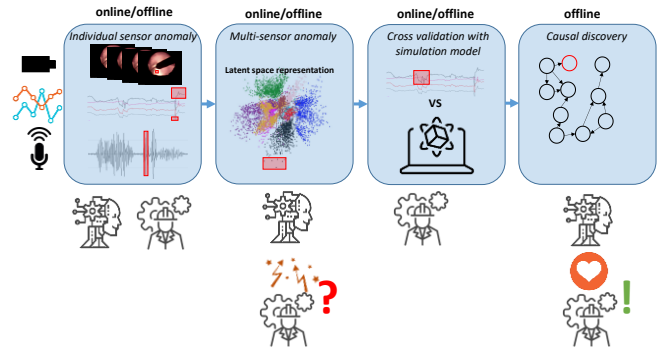

Figure 1. Multi-level data validation framework for multimodal multivariate time series and associated human/machine interctions. Online vs offline level for application respectively to live data and historical data.

<u>Level I - Sensor maintenance and calibration</u>: each sensor must be properly calibrated and a maintenance procedure is defined based on the recommendation of the sensor supplier for the given environment. AI is potentially involved for optimal maintenance scheduling by prescribing actions learnt from historical data to be tackled to prolong the lifetime of sensors.

<u>Level II - Process Min/Max on individual sensor signals</u>: The first level of anomaly detection is provided by the definition of boundary values for the amplitude of each signal as defined by the process expert. Those upper and lower limits are static therefore independent to the process operation.

Level III - Anomaly detection on individual sensor signal: machine learning models are applied independently to the signal of each sensor in order to detect abnormal measurements. Supervised and unsupervised anomaly detection models are providing efficient solutions to identify respectively known anomalies defined by a given pattern, or deviation from a learnt representation of the normality such as unusual amplitude values or spectral information for a given context. Those solutions provide a low-level contextual anomaly detection, as the context is defined solely by the signal itself. Given the simple context, justification of the anomalies can be provided by simple rules or by model interpretability methods such as VAE-LIME [7], an adapted version of the algorithm for Local Interpretable Model-agnostic Explanations (LIME), particularly well adapted to process data having complex inter-variables relations.

Level IV - Anomaly detection on multi-sensor signals: at this level, a representation of the multimodal temporal data combining images, process related multivariate time series, sounds or text, brings a context awareness enabling high-level contextual anomaly detection. Unsupervised deep learning is a solution investigated in the scientific literature for learning abstract representations of complex data allowing to identify causal relations between multimodal inputs, that are dynamic and depending on the process operational mode. Additional prior information such as the sensor type, location (physical on one asset or logical as associated to a sub-process) are defining prior context adding new dimension for learning an enhanced context representation. It is important to separate multimodal time series approaches for anomaly detection and anomaly reasoning. An anomaly detection is characterized by a deviation from the normal relation between sensors signals learnt by an unsupervised deep learning model, while the reasoning of the detection of an anomaly allows the process engineer, for example by means of rules definition, to identify if the deviation is caused by a faulty sensor or by the process itself. The reasoning is crucial, as sensor anomalies must be rejected although rare process phenomena should not and must be tagged specifically for potentially training of a dedicated supervised model such as few-shot learning for predicting their occurrence. Autoencoder (AE) are very popular deep learning networks for learning representation of multivariate time series [8,9], or multimodal time series [10]. In the next section, we present multiple applications of LSTM-based AE for the detection of anomalies in multivariate time series recorded on a blast furnace for the ironmaking industry. The reasoning of an anomaly is a current research subject where scientists are studying the integration of attention mechanism into the architecture of an AE. Some AE architectures have been proposed in the literature with attention mechanism to enhance the learning of better representation to improve the model accuracy but not with the purpose of model reasoning [11,12,13]. A recent article [14] has investigated multiple applications of AE based on its high potential for representation learning, and is exploring the current challenges of their explainability. In [15] an explainability of the reconstruction provided by the AE is given by computing the gradient of the reconstruction error as a measurement of the contribution of each input to that error. Another strategy proposed in [16] consists in restricting the operations of neurons to logical operators such as AND/OR in order to learn features defined by logical combination of inputs. The ultimate challenge of data validation by bridging data anomalies and their causality with the world is enabled by causal discovery as discussed in a next paragraph.

Level V – Cross-validation sensor signal with simulation models results: the availability of simulation models of the underlying process provides means to validate the sensor signals by comparing measurements and simulation results. The validation is however limited to the operational modes of the process respecting the hypothesis inherent to the simulation model. A simulation model is an approximated ideal mathematical representation of a process and gathers the knowledge of experts up to some extent as defined by the hypothesis of the mathematical expression. This level of data validation provides a supervised approach for sensor signal validation, however expertise in the mathematical model and underlying hypothesis is mandatory to judge if any deviations measured between the sensor signals and the model are related to an anomaly on the sensor. Furthermore, understanding a deviation to be the cause of an unrealistic assumption of the model is bringing high value to the process engineer to augment the complexity of that mathematical model.

Level VI - Causal discovery for training dataset validation: The data selected for training a model is required to have a distribution as close as possible to the real world to train a model that is generic, unbiased and therefore robust to the world where predictions will be generated. Extracting the causal relations from observational data is giving relevant information to the domain expert to validate a dataset for training purposes as well as to learn unknown causal relations allowing him to optimize accordingly the underlying process. The outcome of the training dataset validation by the domain expert is leading to the selection of the appropriate architecture to train a model and permits to define requirements to potentially apply transfer learning between multiple domains [17], or to augment the dataset with simulated data to compensate for any known missing causal relations.

Causal discovery helps scientists to interpret the data by defining and testing hypotheses and therefore learn the world to develop better models. Causality is the core of any human judgement and decision allowing to generate explanations and define best actions targeting an improvement to reach the optimal solution in a given environment [18].

Currently, the literature proposes two well-established frameworks with solid mathematical basements for causal inference: *Structural Causal Models* [19] developed by Judea Pearl and based on directed acyclic graphs, and the *Rubin Causal Model* [20] formulated and developed by Donald Rubin and originally proposed by Jerzy Neyman, and based on the contrast of potential outcomes *Y1* and *Y2* caused respectively by *X* and *not X*. In [21], J. Pearl is presenting a three-level causal hierarchy aiming at answering questions such as 'What is?' (level 1 – seeing), 'What if?' (level 2 – doing intervention) and 'what if I had acted differently?' (level 3 – imagining, retrospective). Level 2 and 3 require causal model of the environment. The absence of causal model leads to three fundamental obstacles for the further development of AI: adaptability or robustness to new circumstances (the out-of-distribution representation challenge), explainability of black-box data-driven model and the learning of cause-effect connections for training models to answer level 2 or level 3 questions. Reinforcement learning models are covering, to some extent, the level 2 of causality from learning the world with observational data, experimental data or by means of a simulation model.

To discover causation, often experiments or interventions are required to understand all influencing factors of the target variable by generating *experimental data* for A/B testing for example. This has the drawback of being costly and time consuming or impossible in real life. Challenges for causal discovery lies in the existence of confounders, potentially not observed or even unknown, that introduce correlations between two variables although no causal relationship exists. The enormous quantities of *observational data* recorded for many applications allow scientists to do *causal discovery* (causal graph between variables defined from observational data) or *causal inference* (extrapolation of causal graph – ex: simulate the effect of interventions) [22]. Machine learning models such as neural networks or decision trees are commonly trained to learn correlation relations but this does not imply causation that is usually an asymmetrical relation on the contrary to correlation that is symmetrical [23]. Learning robust relationships such as causation ensures a model to have a precise representation of the world therefore to make better predictions and to reason about events and how they are influenced by outside manipulations [24]. In [25] a proposed approach is a meta-learning of the cause-effect relations for faster adaptation to out-of-distribution introduced by a change of actions of agents (interventions) but not by a change of concept leading to slower adaptation. In [26], an article from the same authors, short-term vs long-term characteristics of the data generation mechanism are learned by a neural network that is parametrized into respectively fast parameters (fast adaptation to interventions) and slow parameters.

Few approaches have been investigated for multivariate time series observational data and are often limited to data assumptions such as stationarity, linearity, absence of noise and confounders [27,28,29,30]. In [31] a Temporal Causal Discovery Framework (TCDF) implementing an attention-based convolutional deep leaning architecture for multivariate time series is proposed as a solution to learn temporal causal graphs including confounders and instantaneous effects as well as temporal delay between a cause and its effect. A survey of methods for temporal causal discovery is defined in [31] where methods are classified according to multiple dimensions depending on features of the methods and assumption on the input data.

## III. RESULTS

The proposed framework has been validated on real world multivariate time series data. The validation on multimodal datasets as well as causal discovery are ongoing research. The multivariate time series data has been recorded on a blast furnace equipped with several thousands of sensors measuring temperatures, pressures, flows, chemical contents, etc. The underlying process of the blast furnace is causing temporal shifts between the time series due to its high inertia. Those temporal shifts are dynamic as they are depending on the operation of the furnace itself, which increases the complexity for the research of a causal discovery approach to implement the Level VI of data validation.

LSTM AE are proven to be a reliable approach for learning the complex natural relations between multivariate time series and therefore detect any deviation from that representation of the normality, leading to warning regarding the data quality in the context given by other time series. LSTM AE is a solution to cover the requirement of Level IV. Fig. 2 illustrates the architecture and approach for unsupervised anomaly detection by differentiating each input time series with its reconstruction provided by the LSTM AE.

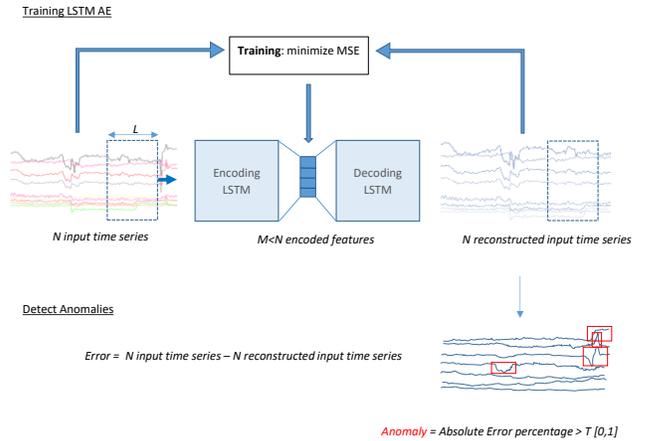

Figure 2. LSTM AE training by minimizing Mean Square Error (MSE) between input and reconstruction, and implementation for multivariate time series anomaly detection.

An application of LSTM AE is the detection of sensor anomaly in the hearth of the blast furnace where multiple thermocouples installed in the refractories of the blast furnace are measuring the temperature (Fig. 3).

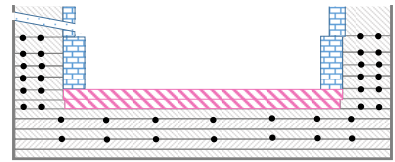

Figure 3. Schematic illustration of thermal hearth sensors location in the wall refractories of a blast furncace (vertical section bottom of the furnace).

Contextual anomalies have been simulated to illustrate the best capability of LSTM AE. First, a loss of signal for one sensor is simulated in Fig. 4 by removing part of the signal on a test dataset and comparing the reconstruction of that signal with its actual value. The missing signal has been reconstructed by the LSTM AE from the learnt correlation with other signals in a training dataset. Another anomaly is generated artificially to simulate a modification of trend of one signal in Fig. 5. The reconstruction of the ideal signal has an absolute error below 4%. The simulated anomaly does not influenced its own reconstruction as LSTM AE has learnt to rely on other sensors for an optimal reconstruction. By applying a threshold on the percentage of reconstruction error, that anomaly can be identified.

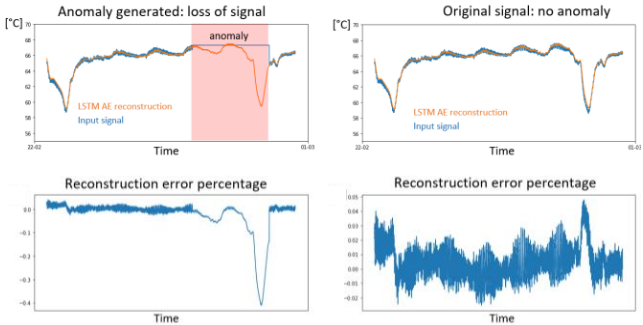

Figure 4. Right: Sensor signal (blue), associated reconstruction (orange), and percentage of reconstruction error for a tested duration of 7 days. Left: Sensor signal (blue) with simulated anomaly (sensor not emitting signal), associated reconstruction and percentage of reconstruction error for the same tested duration.

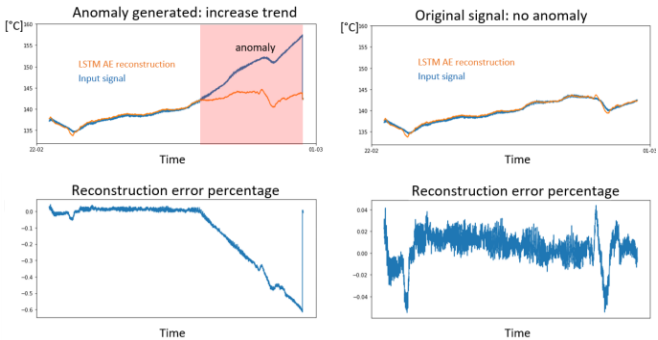

Figure 5. Right: Sensor signal (blue), associated reconstruction (orange), and percentage of reconstruction error for a tested duration of 7 days. Left: Sensor signal (blue) with simulated anomaly (local trend increase of the signal), associated reconstruction (orange) and percentage of reconstruction error for the same tested duration.

Autoencoder reconstruction can be implemented into visualizations in the form of a data validation heatmap as illustrated by Fig. 6, allowing the domain expert to reason from his experience about the occurrence at time *t* of an anomaly and other deviations from the normal relations between time series in a recent past window. The heatmap in Fig. 6 is presenting the variables having a high importance for describing the current thermal state of the blast furnace, and are the input variables of a machine learning model forecasting the hot metal temperature at a time horizon of 3h. The operator is relying on that model to operate the blast furnace and therefore the live input must be validated by advanced methods for multivariate time series as discussed in the section covering the Level IV of the data validation framework.

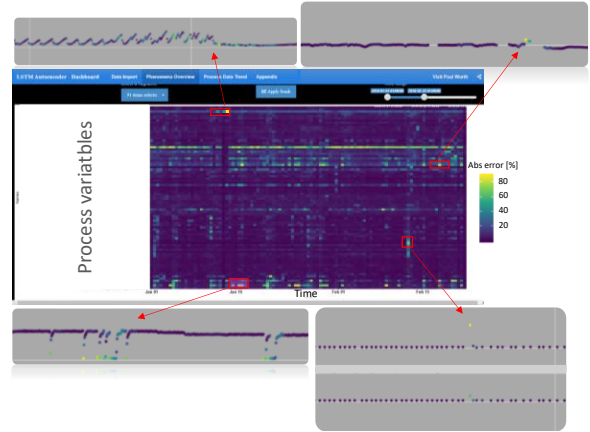

Figure 6. Heatmap based visualization for anomaly monitoring and analysis. A range of two months of input data for hot metal temperature forecasting data-driven model is presented.

IV. CONCLUSION AND PERSPECTIVES

A causal-based data validation framework for multimodal multivariate time series data is presented to identify anomalies in the data by building a stack-based validation with increased level of abstraction for contextual representation, to reach the learning of causal relations being a challenging research subject for the scientific community. Level II and Level III are focusing on the validation of time series data where the context is limited to the time series itself, and where existing state-of-the-art methods are providing a robust solution as a result of a high level of maturity in that research subject. The first challenge lies in Level IV where the context is defined by multiple signals originating from multiple sensors recording multimodal data. Unsupervised deep learning is bringing its power to learn complex abstract representation allowing a disentanglement of the normal relations between multimodal time series, and rare events that can be associated identically to a sensor failure or unexpected process phenomena. The latter case is enabling the process engineer to further understand and therefore optimize his process, but a requirement is the learning of the causal relations of Level VI to identify rare process phenomena from bad sensor measurements. Autoencoders are popular algorithms for complex representation learning. An LSTM AE has been successfully evaluated on the use case of the blast furnace to learn the normal relations between multivariate time series allowing a live measurement of the deviation of the current input time series values from the reference learnt by the algorithm. The last level is dedicated to the offline validation of a training dataset by learning the causal relations between the variables and exposed them to a domain expert in charge of judging their relevance. A dataset for training a data-driven model must be unbiased and statistically representative for the process to model. A data validation integrating the assessment of the process causality from observational data analysis, when anomalies have been identified, corrected or removed, is the main requirement leading to a mature procedure for training unbiased machine learning models, and is the challenge for the next generation of machines.

As a perspective to this work, the framework will be further evaluated by training a multimodal autoencoder combining time series and images to increase further the learning of a larger context. As exposed, some recent papers have emerged to cover this subject and are defining the initial iterations for this research. Another long-term research subject is the learning of causal relations first on multivariate time series and later to extend this to multimodal time series. The combination of the research to cover Level IV and Level VI is leading to a solution for unsupervised Root Cause Analysis built from the basement of learning the causal relations and not solely the correlative information between variables that is known to be incorrectly translated to causal relations with the existence of confounders. However, for the first iterations of the related research, it will be assumed that no confounders are existing in the dataset and therefore associating causality and correlation by applying attention mechanisms for multivariate time series. This will serve as references on a benchmark of multiple solutions for causality discovery in multimodal multivariate time series.

REFERENCES

[1] C. Schockaert, F. Hansen, F. Giroldini, A. Schmitz, "Blast Furnace Data Validation Using Deep Learning as an Enabler for Autonomous Blast Furnace-CedricSchockaert", in press, AISTech 2020, Cleveland, USA, 2020.

[2] Y. Bengio, "From System 1 Deep Learning to System 2 Deep Learning", NIPS, Toronto, 2019.

[3] J. Ding, X.-H. Hu and V. N. Gudivada, "A Machine Learning Based Framework for Verification and Validation of Massive Scale Image Data." IEEE Transactions on Big Data, 2017.

[4] H. Foidl and M. Felderer, "Risk-based data validation in machine learning-based software systems," In Proceedings of the 3rd ACM SIGSOFT International Workshop on Machine Learning Techniques for Software Quality Evaluation, New York, USA, 2019, pp. 13-18.

[5] S. Schelter, D. Lange, P. Schmidt, M. Celikel, F. Biessmann, and A. Grafberger, "Automating Large-Scale Data Quality Verification," Proceedings of the VLDB Endowment 11, 12, 2018, pp. 1781-1794.

[6] D. Sculley, G. Holt, D. Golovin, E. Davydov, T. Phillips, D. Ebner, V. Chaudhary, M. Young, J.-F. Crespo, D. Dennison, " Hidden Technical Debt in Machine Learning Systems," In NIPS. MIT Press, Cambridge, MA, USA, 2015, pp. 2503-2511.

[7] C. Schockaert, V. Macher, A. Schmitz, "VAE-LIME: Deep Generative Model Based Approach for Local Data-Driven Model Interpretability Applied to the Ironmaking Industry," preprint 2020, DOI:10.13140/RG.2.2.15447.70567.

[8] F. M. Bianchi, L. Livi, K. Øyvind Mikalsen, M. Kampffmeyer, R. Jenssen, "Learning representations for multivariate time series with missing data using Temporal Kernelized Autoencoders," arXiv 2018, arXiv: 1805.03473.

[9] A. Sagheer, M. Kotb, "Unsupervised Pre-training of a Deep LSTM-based Stacked Autoencoder for Multivariate Time Series Forecasting Problems," Sci Rep 9, 19038, 2019.

[10] X. Yang, P. Ramesh, R. Chitta, S. Madhvanath, E. A. Bernal, J. Luo, "Deep Multimodal Representation Learning from Temporal Data," arXiv 2019, arXiv: 1704.03152.

[11] J. Pereira and M. Silveira, "Unsupervised Anomaly Detection in Energy Time Series Data Using Variational Recurrent Autoencoders with Attention," 17th IEEE International Conference on Machine Learning and Applications (ICMLA), Orlando, FL, 2018, pp. 1275-1282.

[12] T. Sun and A. A. Wu, "Sparse Autoencoder with Attention Mechanism for Speech Emotion Recognition," IEEE International Conference on Artificial Intelligence Circuits and Systems (AICAS), Hsinchu, Taiwan, 2019, pp. 146-149.

[13] T. Tian, Z. Fang, "Attention-based Autoencoder Topic Model for Short Texts," Procedia Computer Science, 151, 2019, pp.1134-1139.

[14] D. Charte, F. Charte, M. J. del Jesus, F. Herrera, "An analysis on the use of autoencoders for representation learning: fundamentals, learning task case studies, explainability and challenges," arXiv 2020, arXiv:2005.10516v1

[15] Q. P. Nguyen, K. W. Lim, D. M. Divakaran, K. H. Low, M. C. Chan, Gee, "A gradient-based explainable variational autoencoder for network anomaly detection," in: 2019 IEEE Conference on Communications and Network Security (CNS), IEEE, 2019, pp. 91-99.

[16] R. Al-Hmouz, W. Pedrycz, A. Balamash, A. Morfeq, "Logic-driven autoencoders," Knowledge-Based Systems 183, 2019, 104874.

[17] C. Schockaert, H. Hoyez, "MTS-CycleGAN: An Adversarial-based Deep Mapping Learning Network for Multivariate Time Series Domain Adaptation Applied to the Ironmaking Industry," preprint 2020, DOI:10.13140/RG.2.2.27191.75680.

[18] D. Danks, The Psychology of Causal Perception and Reasoning. In The Oxford Handbook of Causation; Helen Beebee, C.H., Menzies, P., Eds.; Oxford University Press: Oxford, UK, 2009; Chapter 21, pp. 447-470.

[19] J. Pearl and D Mackenzie, The Book of Why: The New Science of Cause and Effect (1st. ed.). Basic Books, Inc., USA, 2018.

[20] J. Sekhon, "The Neyman-Rubin Model of Causal Inference and Estimation via Matching Methods," Oxford handbook of political methodology, 2008.

[21] J. Pearl, "The seven tools of causal inference, with reflections on machine learning." Communications of the ACM 62, 2019, pp. 54-60.

[22] K. Zhang, B. Schölkopf, P. Spirtes, C. Glymour, "Learning causality and causality-related learning: Some recent progress," Natl. Sci. Rev. 2017, 5, pp. 26-29.

[23] S. Kleinberg, Causality, Probability, and Time; Cambridge University Press: Cambridge, UK, 2013.

[24] P. Spirtes, "Introduction to causal inference," J. Mach. Learn. Res. 2010, 11, pp. 1643-1662.

[25] Y. Bengio, T. Deleu, N. Rahaman, R. Ke, S. Lachapelle, O. Bilaniuk, A. Goyal, C. Pal, "A Meta-Transfer Objective for Learning to Disentangle Causal Mechanisms," arXiv 2019, arXiv: 1901.10912.

[26] R. Ke, O. Bilaniuk, A. Goyal, S. Bauer, H. Larochelle, C. Pal, Y. Bengio, "Learning Neural Causal Models from Unknown Interventions," arXiv 2019, arXiv: 1910.01075.

[27] J. Runge, D. Sejdinovic, S. Flaxman, "Detecting causal associations in large nonlinear time series datasets," arXiv 2017, arXiv:1702.07007.

[28] Y. Huang, S. Kleinberg, "Fast and Accurate Causal Inference from Time Series Data," In Proceedings of the FLAIRS Conference, Hollywood, FL, USA, 18–20 May 2015; pp. 49-54.

[29] M. Hu, H. Liang, "A copula approach to assessing Granger causality," NeuroImage 2014, 100, 125-134.

[30] A. Papana, C. Kyrtsou, D. Kugiumtzis, C. Diks, "Detecting causality in non-stationary time series using partial symbolic transfer entropy: Evidence in financial data," Comput. Econ. 2016, 47, pp. 341-365.

[31] M. Nauta, D. Bucur & C. Seifert, "Causal Discovery with Attention-Based Convolutional Neural Networks," Machine Learning and Knowledge Extraction, 1(1), 2019, pp. 312-340.

AUTHORS' BACKGROUND

| Your Name | Title* | Research Field | Personal website |
|---|---|---|---|
| Cedric Schockaert | Dr. | Autonomous blast furnace:<br>1. Unsupervised deep learning for representation learning applied to multimodal and multivariate time series<br>2. Model interpretability<br>3. Transfer learning for multivariate time series<br>4. Deep reinforcement learning | |